\documentclass[sigconf]{acmart} 
\usepackage{graphicx}
\usepackage{amsmath}
\usepackage{subcaption}
\usepackage{epstopdf}
\usepackage{amsthm}
\usepackage{booktabs}
\usepackage{hyperref}
\usepackage{algorithm}
\usepackage{algorithmic}
\usepackage{multirow}
\usepackage{dsfont}

\newcommand{\RNum}[1]{\uppercase\expandafter{\romannumeral #1\relax}}
 
\setcopyright{none}

\acmDOI{00.000/000_0}
 
\acmISBN{000-0000-00-000/00/00}
 
\acmConference[0000]{000}{0000}{Beijing, China} 
\acmYear{0000}
\copyrightyear{0000}
\acmPrice{00.00}

\begin{document}
	\title{Clustering with Neural Network and Index}
	
	\author{Gangli Liu}
	\affiliation{%
		\institution{Tsinghua University}
	}
	\email{gl-liu13@mails.tsinghua.edu.cn}

	\begin{abstract}
A new model called \emph{C}lustering with \emph{N}eural \emph{N}etwork and \emph{I}ndex (CNNI) is introduced. CNNI uses a Neural Network to cluster data points. Training of the Neural Network mimics supervised learning, with an internal clustering evaluation index acting as the loss function. An experiment is conducted to test the feasibility of the new model, and compared with results of other clustering models like K-means and Gaussian Mixture Model (GMM). The result shows CNNI can work properly for clustering data; CNNI equipped with MMJ-SC, achieves the first parametric (inductive) clustering model that can deal with non-convex shaped (non-flat geometry) data.

\end{abstract}
 
	\maketitle
 
\section{Introduction}

Cluster analysis is the grouping of objects such that objects in the same cluster are more similar to each other than they are to objects in another cluster. Clustering is a type of unsupervised learning method of machine learning. In unsupervised learning, the inferences are drawn from the data sets which do not contain labeled output variable.  Due to its intensive applications in data analysis, people have developed a wide variety of  clustering algorithms. Such as K-Means, Gaussian mixtures, Spectral clustering \cite{ng2001spectral}, Hierarchical clustering \cite{johnson1967hierarchical}, DBSCAN \cite{schubert2017dbscan}, BIRCH \cite{zhang1996birch} etc. 

Evaluation (or ``validation") of clustering results is as difficult as the clustering itself \cite{pfitzner2009characterization}. 
Popular approaches involve ``internal" evaluation and  ``external" evaluation.  In internal evaluation, a clustering result is evaluated based on the data that was clustered itself. Popular internal evaluation indices are Davies-Bouldin index \cite{ petrovic2006comparison}, Silhouette coefficient \cite{aranganayagi2007clustering}, Dunn index \cite{bezdek1995cluster}, and Calinski-Harabasz index  \cite{maulik2002performance} etc.   In external evaluation, the clustering result is compared to an existing ``ground truth" classification, such as the Rand index \cite{yeung2001details}.  

There have been proposed several types of Artificial Neural Networks (ANNs) with numerous different implementations for clustering tasks. Most of these Neural Networks apply competitive learning rather than error-correction learning as most other types of Neural Networks do. 

Unlike Neural Networks used in competitive learning, CNNI mimics an error-correction learning model.

	\begin{figure*} 
	\begin{subfigure}{0.3\textwidth}
		\includegraphics[width=\linewidth]{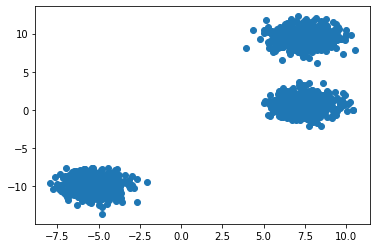}
		\caption{data 0}  
	\end{subfigure}    
	\begin{subfigure}{0.3\textwidth}
		\includegraphics[width=\linewidth]{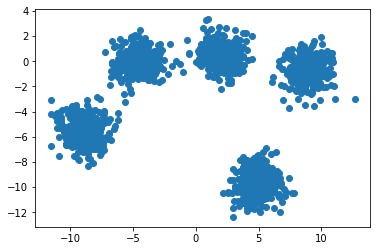}
		\caption{data 1}  
	\end{subfigure}    
	\begin{subfigure}{0.3\textwidth}
	\includegraphics[width=\linewidth]{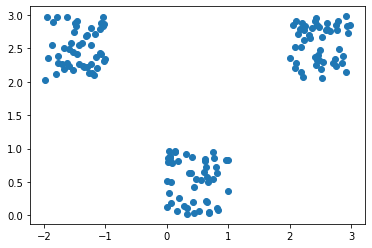}
	\caption{data 2}  
\end{subfigure}    

	\caption{Three datasets for Experiment \RNum{1}} \label{fig:three-data}
\end{figure*}
	\begin{figure*} 
	\begin{subfigure}{0.24\textwidth}
		\includegraphics[width=\linewidth]{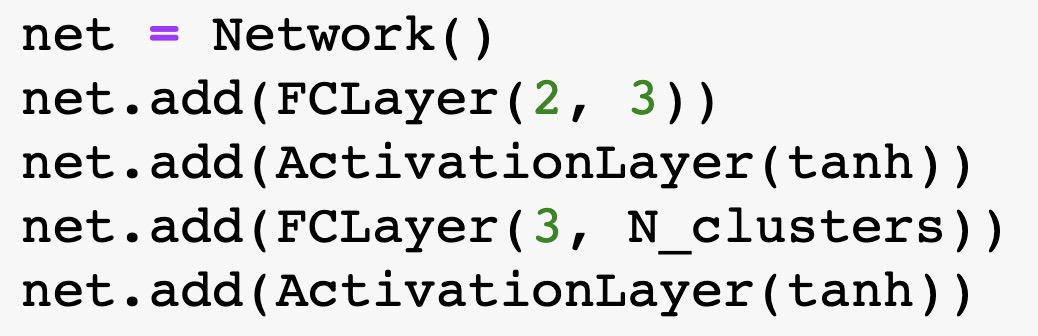}
		\caption{NN A}  
	\end{subfigure}    
	\begin{subfigure}{0.24\textwidth}
		\includegraphics[width=\linewidth]{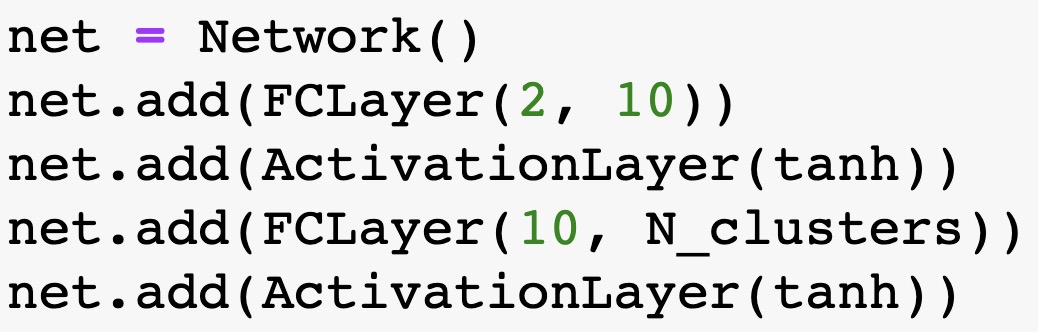}
		\caption{NN B}  
	\end{subfigure}    
	\begin{subfigure}{0.24\textwidth}
	\includegraphics[width=\linewidth]{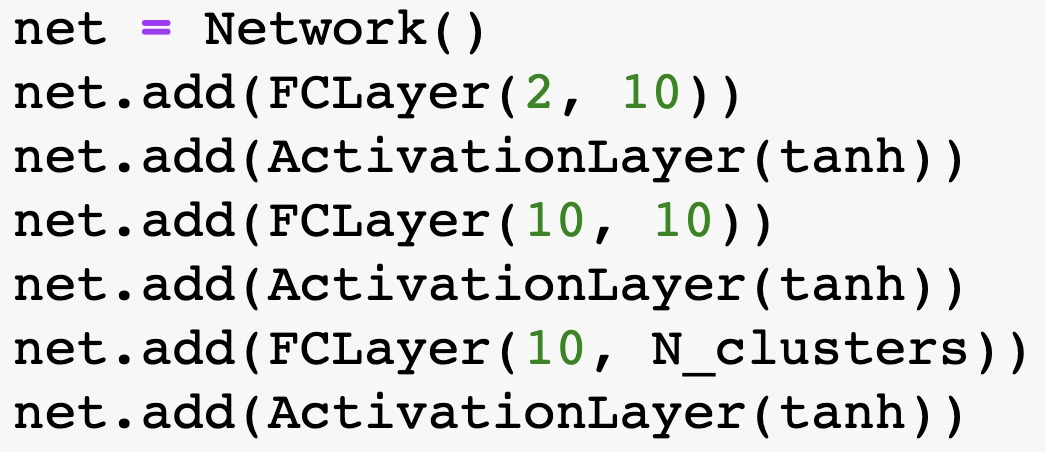}
	\caption{NN C}  
\end{subfigure}    
	\begin{subfigure}{0.24\textwidth}
	\includegraphics[width=\linewidth]{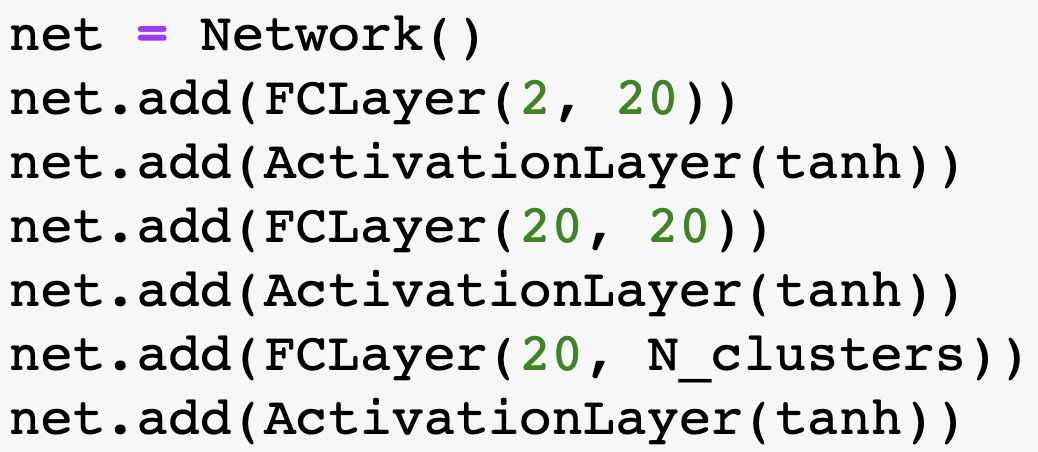}
	\caption{NN D}  
\end{subfigure} 
	\caption{Four Neural Networks} \label{fig:three-nn}
\end{figure*}

\section{RELATED WORK} \label{Sec_related}

\subsection{Internal clustering evaluation index}

Calinski-Harabasz index \cite{maulik2002performance}, Silhouette coefficient \cite{aranganayagi2007clustering}, and Davies-Bouldin index \cite{ petrovic2006comparison} are three of the most popular techniques for internal clustering evaluation. 

Besides the three popular ones, numerous criteria have been proposed in the literature \cite{vendramin2010relative,chou2004new,vzalik2011validity,baya2013many,moulavi2014density}. Such as DBCV \cite{moulavi2014density}, Xie-Beni (XB) \cite{xie1991validity}, CDbw \cite{halkidi2008density}, S\_Dbw \cite{halkidi2001clustering}, and RMSSTD \cite{halkidi2001clustering2001}. Besides, new cluster validity indices keep emerging, such as the CVNN \cite{liu2013understanding}, CVDD \cite{hu2019internal}, DSI \cite{guan2020internal}, SCV \cite{xu2020efficient}, AWCD \cite{li2020new} and VIASCKDE \cite{csenol2022viasckde}. The author also proposed one \cite{liu2022new}.

Because only Silhouette coefficient is used in this paper, following is some detailed recount about it.

\subsection{Silhouette coefficient (SC)}

The Silhouette coefficient for a single sample is given as:
\begin{equation*}
s = \frac{b - a}{max(a, b)}
\label{eq:S_index}
\end{equation*}
where $ a $ is the mean distance between a sample and all other points in the same class. $ b $ is the mean distance between a sample and all other points in the next nearest cluster. The Silhouette coefficient for a set of samples is given as the mean of Silhouette coefficient for each sample. The range of Silhouette coefficient is  $ [-1, 1]$. A higher Silhouette coefficient score relates to a model with better defined clusters.

\subsection{Artificial Neural Network}
Artificial Neural Networks (ANNs), usually simply called Neural Networks (NNs) or neural nets, are based on a collection of connected units or nodes called artificial neurons, which loosely model the neurons in a biological brain \cite{anderson1995introduction,gurney2018introduction}. Typically, neurons are aggregated into layers. Different layers may perform different transformations on their inputs. Signals travel from the first layer (the input layer), to the last layer (the output layer), possibly after traversing the layers multiple times.

Training of a Neural Network is usually conducted by determining the difference between the processed output of the network (often a prediction) and a target output \cite{lu2022learning}. This difference is the error. The network then adjusts its parameters (weights and biases) according to a learning rule and using this error value. Successive adjustments will cause the Neural Network to produce output which is increasingly similar to the target output. After a sufficient number of these adjustments the training can be terminated based upon certain criteria. This is known as supervised learning.

\subsection{Self-Organizing Map}
Probably, the most popular type of neural nets used for clustering is called Self-Organizing Map (SOM), also known as Kohonen network \cite{kohonen1990self}. ``There are many different types of Kohonen networks. These Neural Networks are very different from most types of Neural Networks used for supervised tasks. Kohonen networks consist of only two layers". The structure of SOM may look like perceptrons. ``However, SOM works in a different way than perceptrons or any other networks for supervised learning".

In a SOM, each neuron of an output layer holds a vector whose dimensionality equals the number of neurons in the input layer. ``In turn, the number of neurons in the input layer must be equal to the dimensionality of data points given to the network". 

``When the network gets an input, the input is traversed into only one neuron of the output layer, whose value is closer to the input vector than the values of other output neurons. This neuron is called a winning neuron or the best matching unit (BMU). This is a very important distinction from many other types of Neural Networks, in which values propagate to all neurons in a succeeding layer. And this process constitutes the principle of competitive learning". \footnote{\url{http://www.kovera.org/neural-network-for-clustering-in-python/}}

CNNI is also a Neural Network for clustering, but it works like Neural Networks for supervised learning. The only difference is that we use an internal clustering evaluation index as the loss function,  adjust the weights of the Neural Network to reduce the loss (improve the value of the index).

\subsection{Deep Clustering}
Deep clustering models use a Deep Neural Network (DNN) to transform high dimensional data into more clustering-friendly representations without manual feature extraction, then use a classic clustering model to cluster the transformed data \cite{aljalbout2018clustering,ren2022deep,zhou2022comprehensive}. In these models, a Neural Network is used as a helper for clustering, the ``clustering" part of these models is still accomplished by classic clustering models, such as K-means, Agglomerative clustering etc. In CNNI, no other classic clustering models are needed.

\section{ The Structure of CNNI} 
The structure of CNNI is simple: a Neural Network for supervised learning plus an internal clustering evaluation index. The index acts as the loss function, because there is no target output associated with each input data point in clustering scenario. 

The number of neurons in the input layer of CNNI equals to the dimension of the data points given to the network. The number of neurons in the output layer of CNNI equals to number of $ K $ clusters we want to classify. By comparing values of output neurons, label of one data point is obtained (e.g., find out the maximum of output neurons).

Training of CNNI has some difference from other supervised learning Neural Networks. We need to compute each data point's label according to the Neural Network's current state, then calculate the value of the clustering evaluation index, according to the labels of all data points.  Adjustment of the weights of the Neural Network is based on the value of the index.

	\begin{figure*} 
	\begin{subfigure}{0.19\textwidth}
		\includegraphics[width=\linewidth]{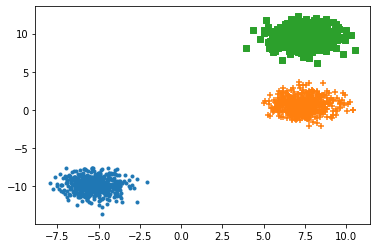}
		\caption{NN A}  
	\end{subfigure}    
	\begin{subfigure}{0.19\textwidth}
		\includegraphics[width=\linewidth]{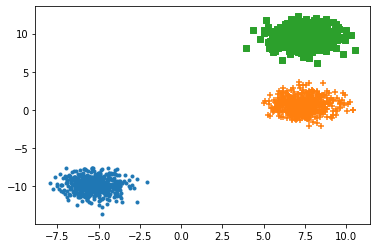}
		\caption{NN B}  
	\end{subfigure}    
	\begin{subfigure}{0.19\textwidth}
	\includegraphics[width=\linewidth]{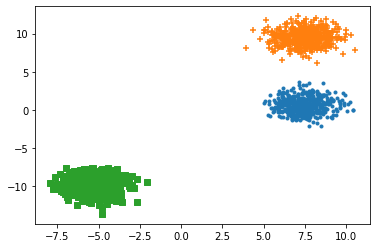}
	\caption{NN C}  
\end{subfigure}    
\begin{subfigure}{0.19\textwidth}
\includegraphics[width=\linewidth]{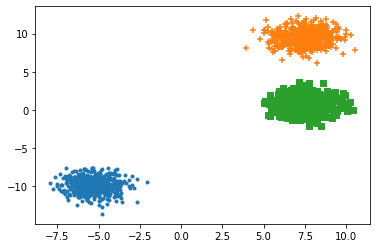}
\caption{K-means}  
\end{subfigure}    
\begin{subfigure}{0.19\textwidth}
\includegraphics[width=\linewidth]{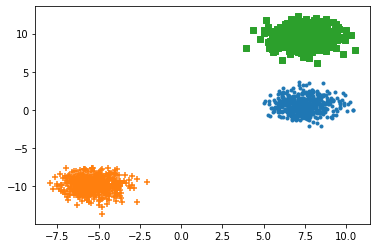}
\caption{GMM}  
\end{subfigure}  

	\begin{subfigure}{0.19\textwidth}
	\includegraphics[width=\linewidth]{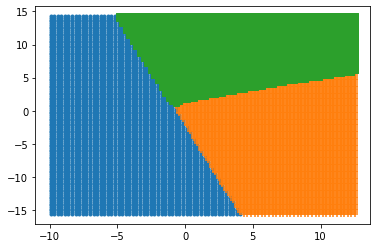}
	\caption{NN A}  
\end{subfigure}    
\begin{subfigure}{0.19\textwidth}
	\includegraphics[width=\linewidth]{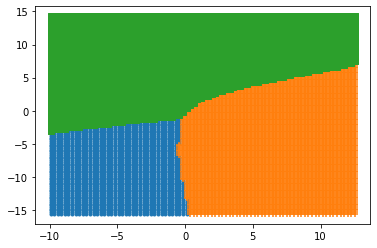}
	\caption{NN B}  
\end{subfigure}    
\begin{subfigure}{0.19\textwidth}
	\includegraphics[width=\linewidth]{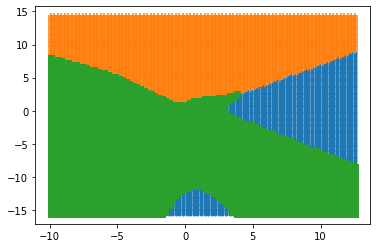}
	\caption{NN C}  
\end{subfigure}    
\begin{subfigure}{0.19\textwidth}
	\includegraphics[width=\linewidth]{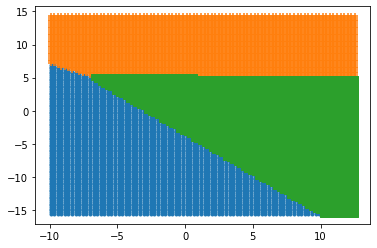}
	\caption{K-means}  
\end{subfigure}    
\begin{subfigure}{0.19\textwidth}
	\includegraphics[width=\linewidth]{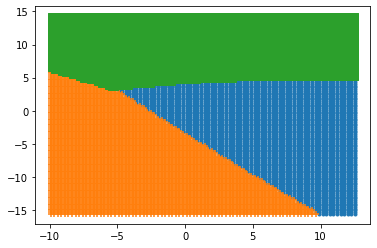}
	\caption{GMM}  
\end{subfigure}

	\caption{Clustering results and decision boundaries of data 0} \label{fig:resdata0}
\end{figure*}
	\begin{figure*} 
	\begin{subfigure}{0.19\textwidth}
		\includegraphics[width=\linewidth]{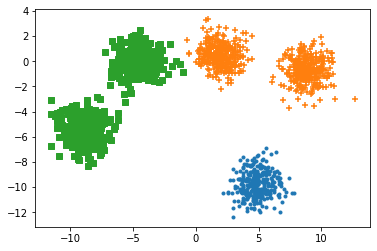}
		\caption{NN A}  
	\end{subfigure}    
	\begin{subfigure}{0.19\textwidth}
		\includegraphics[width=\linewidth]{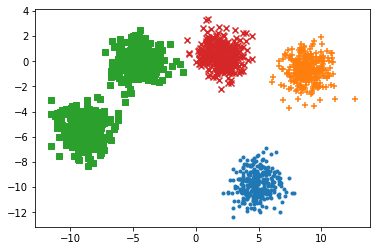}
		\caption{NN B}  
	\end{subfigure}    
	\begin{subfigure}{0.19\textwidth}
		\includegraphics[width=\linewidth]{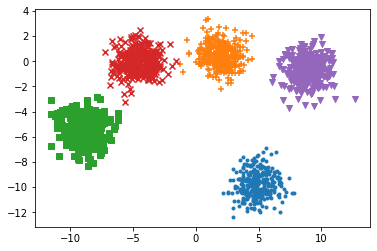}
		\caption{NN C}  
	\end{subfigure}    
	\begin{subfigure}{0.19\textwidth}
		\includegraphics[width=\linewidth]{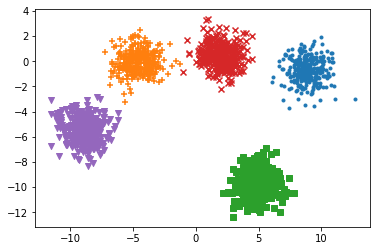}
		\caption{K-means}  
	\end{subfigure}    
	\begin{subfigure}{0.19\textwidth}
		\includegraphics[width=\linewidth]{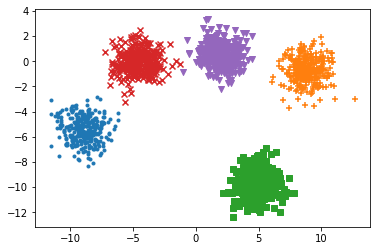}
		\caption{GMM}  
	\end{subfigure}  
	
	\begin{subfigure}{0.19\textwidth}
		\includegraphics[width=\linewidth]{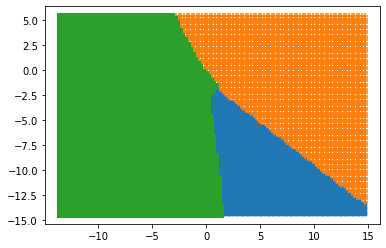}
		\caption{NN A}  
	\end{subfigure}    
	\begin{subfigure}{0.19\textwidth}
		\includegraphics[width=\linewidth]{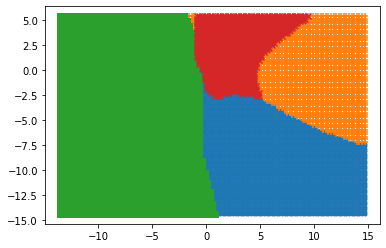}
		\caption{NN B}  
	\end{subfigure}    
	\begin{subfigure}{0.19\textwidth}
		\includegraphics[width=\linewidth]{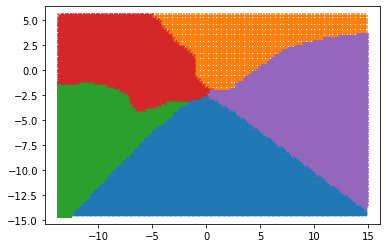}
		\caption{NN C}  
	\end{subfigure}    
	\begin{subfigure}{0.19\textwidth}
		\includegraphics[width=\linewidth]{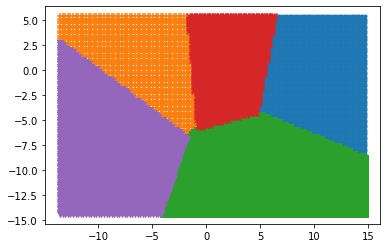}
		\caption{K-means}  
	\end{subfigure}    
	\begin{subfigure}{0.19\textwidth}
		\includegraphics[width=\linewidth]{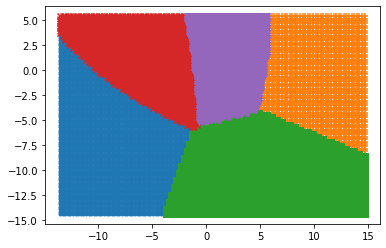}
		\caption{GMM}  
	\end{subfigure}
	
	\caption{Clustering results and decision boundaries of data 1} \label{fig:resdata1}
\end{figure*}
	\begin{figure*} 
	\begin{subfigure}{0.19\textwidth}
		\includegraphics[width=\linewidth]{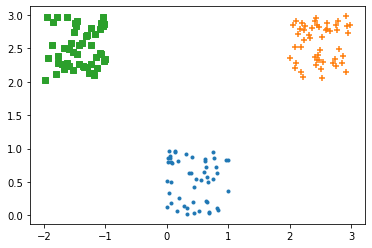}
		\caption{NN A}  
	\end{subfigure}    
	\begin{subfigure}{0.19\textwidth}
		\includegraphics[width=\linewidth]{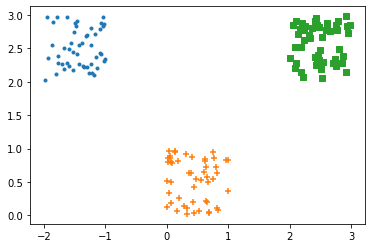}
		\caption{NN B}  
	\end{subfigure}    
	\begin{subfigure}{0.19\textwidth}
		\includegraphics[width=\linewidth]{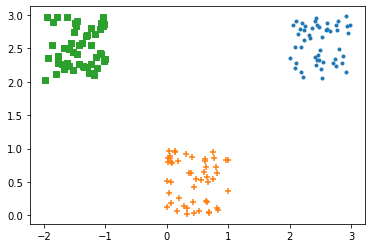}
		\caption{NN C}  
	\end{subfigure}    
	\begin{subfigure}{0.19\textwidth}
		\includegraphics[width=\linewidth]{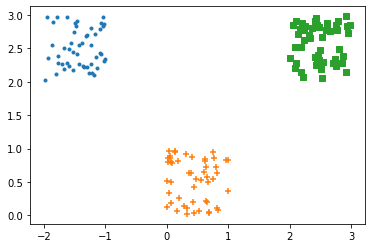}
		\caption{K-means}  
	\end{subfigure}    
	\begin{subfigure}{0.19\textwidth}
		\includegraphics[width=\linewidth]{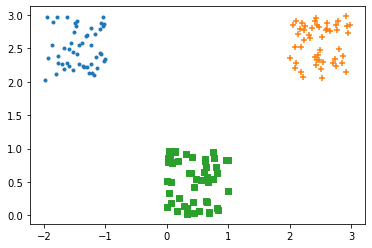}
		\caption{GMM}  
	\end{subfigure}  
	
	\begin{subfigure}{0.19\textwidth}
		\includegraphics[width=\linewidth]{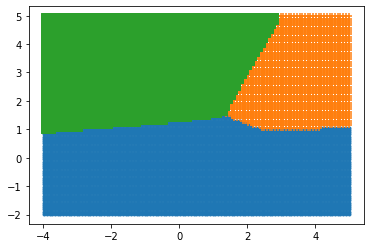}
		\caption{NN A}  
	\end{subfigure}    
	\begin{subfigure}{0.19\textwidth}
		\includegraphics[width=\linewidth]{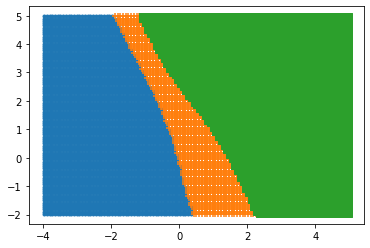}
		\caption{NN B}  
	\end{subfigure}    
	\begin{subfigure}{0.19\textwidth}
		\includegraphics[width=\linewidth]{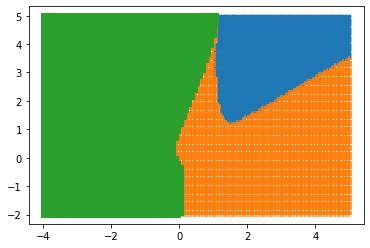}
		\caption{NN C}  
	\end{subfigure}    
	\begin{subfigure}{0.19\textwidth}
		\includegraphics[width=\linewidth]{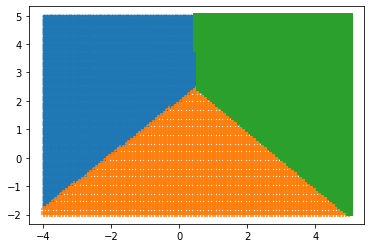}
		\caption{K-means}  
	\end{subfigure}    
	\begin{subfigure}{0.19\textwidth}
		\includegraphics[width=\linewidth]{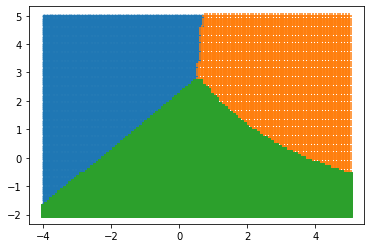}
		\caption{GMM}  
	\end{subfigure}
	
	\caption{Clustering results and decision boundaries of data 2} \label{fig:resdata2}
\end{figure*}
	\begin{figure*} 
	\begin{subfigure}{0.19\textwidth}
		\includegraphics[width=\linewidth]{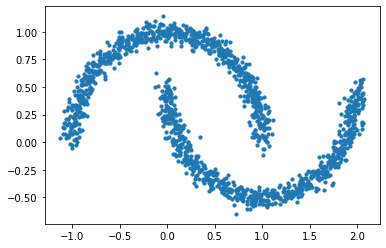}
		\caption{data 3}  
	\end{subfigure}    
	\begin{subfigure}{0.19\textwidth}
		\includegraphics[width=\linewidth]{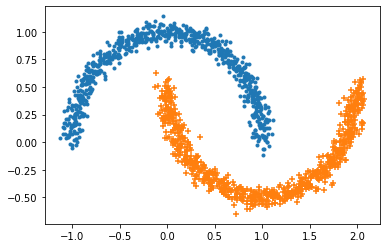}
		\caption{Clu-res}  
	\end{subfigure}    
	\begin{subfigure}{0.19\textwidth}
	\includegraphics[width=\linewidth]{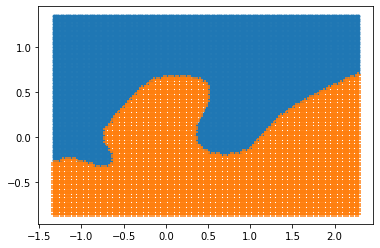}
	\caption{Deci-boun}  
\end{subfigure}    
\begin{subfigure}{0.19\textwidth}
\includegraphics[width=\linewidth]{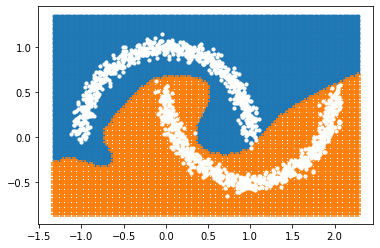}
\caption{Deci-boun-data}  
\end{subfigure}    
\begin{subfigure}{0.19\textwidth}
\includegraphics[width=\linewidth]{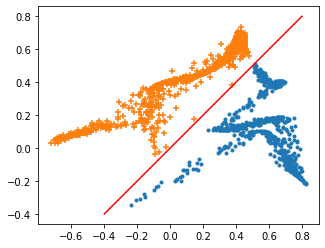}
\caption{Trans-data}  
\end{subfigure}  

	\begin{subfigure}{0.19\textwidth}
	\includegraphics[width=\linewidth]{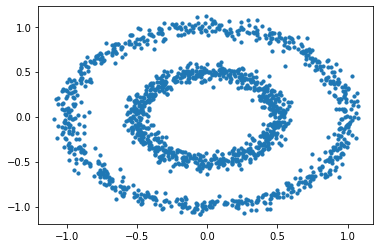}
	\caption{data 4}  
\end{subfigure}    
\begin{subfigure}{0.19\textwidth}
	\includegraphics[width=\linewidth]{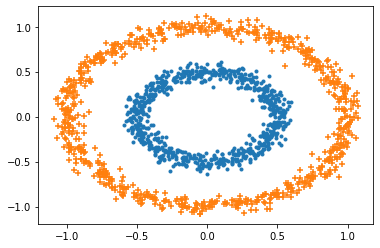}
	\caption{Clu-res}  
\end{subfigure}    
\begin{subfigure}{0.19\textwidth}
	\includegraphics[width=\linewidth]{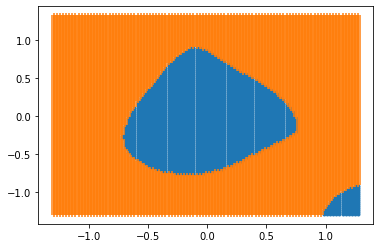}
	\caption{Deci-boun}  
\end{subfigure}    
\begin{subfigure}{0.19\textwidth}
	\includegraphics[width=\linewidth]{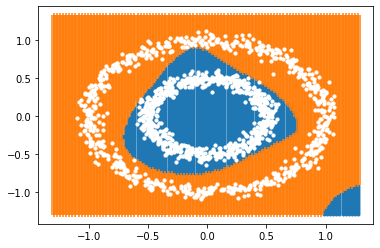}
	\caption{Deci-boun-data}  
\end{subfigure}    
\begin{subfigure}{0.19\textwidth}
	\includegraphics[width=\linewidth]{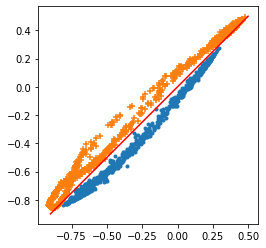}
	\caption{Trans-data}  
\end{subfigure}  

	\begin{subfigure}{0.19\textwidth}
	\includegraphics[width=\linewidth]{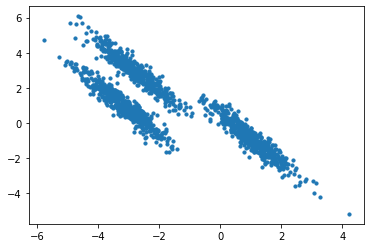}
	\caption{data 5}  
\end{subfigure}    
\begin{subfigure}{0.19\textwidth}
	\includegraphics[width=\linewidth]{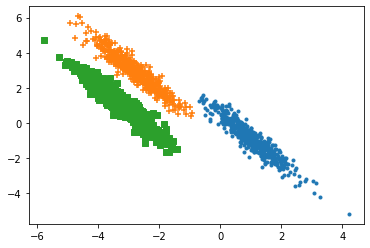}
	\caption{Clu-res}  
\end{subfigure}    
\begin{subfigure}{0.19\textwidth}
	\includegraphics[width=\linewidth]{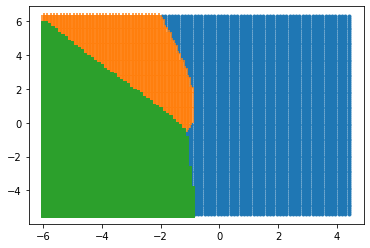}
	\caption{Deci-boun}  
\end{subfigure}    
\begin{subfigure}{0.19\textwidth}
	\includegraphics[width=\linewidth]{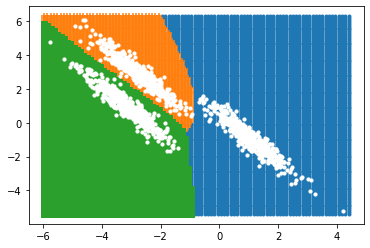}
	\caption{Deci-boun-data}  
\end{subfigure}    
\begin{subfigure}{0.19\textwidth}
	\includegraphics[width=\linewidth]{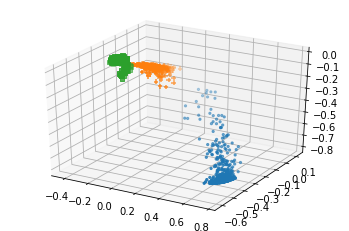}
	\caption{Trans-data}  
\end{subfigure}

	\caption{ Clustering results, decision boundaries, and transformed data of data 3,4,5} \label{fig:line_sep}	

\end{figure*}
	\begin{figure*} 
	\includegraphics[width=0.9\textwidth]{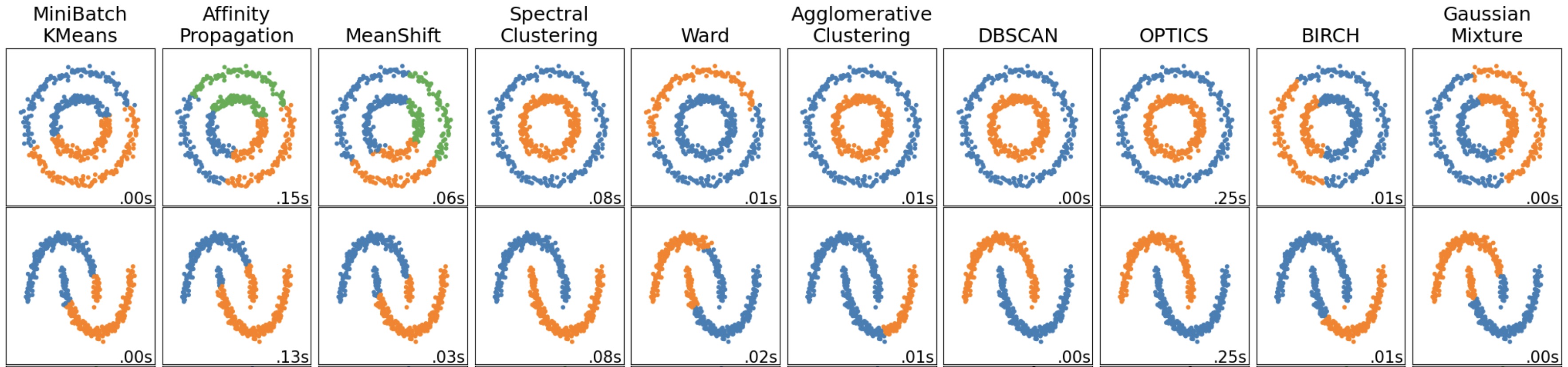}
	\caption{Clustering results of two-moons and two-circles by other models (from the scikit-learn project) } 
	\label{fig:other_clu}	

\end{figure*}
	\begin{figure*} 
	\begin{subfigure}{0.25\textwidth}
		\includegraphics[width=\linewidth]{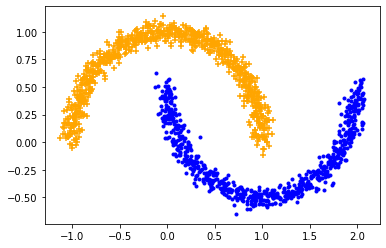}
		\caption{Clu-res}  
	\end{subfigure}    
	\begin{subfigure}{0.25\textwidth}
		\includegraphics[width=\linewidth]{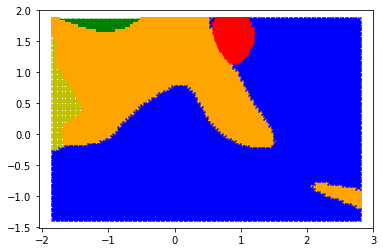}
		\caption{Deci-boun}  
	\end{subfigure}    
	\begin{subfigure}{0.25\textwidth}
	\includegraphics[width=\linewidth]{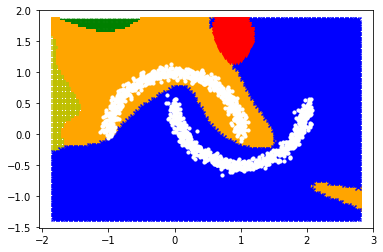}
	\caption{Deci-boun-data}  
\end{subfigure}    

	\begin{subfigure}{0.25\textwidth}
	\includegraphics[width=\linewidth]{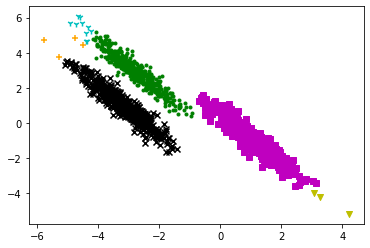}
	\caption{Clu-res}  
\end{subfigure}    
\begin{subfigure}{0.25\textwidth}
	\includegraphics[width=\linewidth]{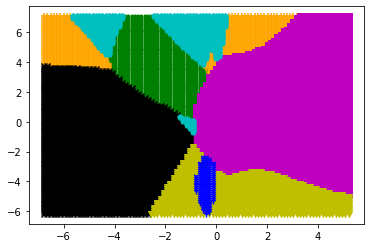}
	\caption{Deci-boun}  
\end{subfigure}    
\begin{subfigure}{0.25\textwidth}
	\includegraphics[width=\linewidth]{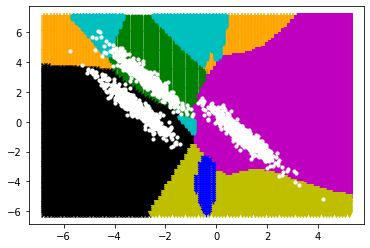}
	\caption{Deci-boun-data}  
\end{subfigure}

	\caption{ Clustering results and decision boundaries of automatic model selection} 
	\label{fig:model_selec}	

\end{figure*}
\section{Experiments} 
We tested the practicability of CNNI  with several experiments. In the experiments, six synthetic datasets are to be clustered by four simple feed-forward Neural Networks (FNNs) with different model capacity (Figure \ref{fig:three-nn}). All the synthetic datasets are generated with library functions from the scikit-learn project \cite{scikit-learn}.

Coordinate descent is used to train the Neural Networks.
 
\subsection{Coordinate descent} 
Coordinate descent is an optimization algorithm that successively minimizes along coordinate directions to find the minimum of a function. At each iteration, the algorithm determines a coordinate or coordinate block via some coordinate selection rule, then exactly or inexactly minimizes over the corresponding coordinate hyperplane while fixing all other coordinates or coordinate blocks. A line search along the coordinate direction can be performed at the current iterate to determine the appropriate step size. Coordinate descent is applicable in both differentiable and derivative-free contexts. \footnote{\url{https://en.wikipedia.org/wiki/Coordinate_descent}}

In the experiments, we use a naive coordinate descent algorithm (Algorithm 1) for adjusting parameters of the Neural Network. People can test more advanced optimization techniques like gradient descent, genetic algorithms, and particle swarm optimization etc.

\begin{algorithm}
	\caption{Naive coordinate descent for CNNI}  
	\begin{algorithmic}[1]  
		\REQUIRE ~~\\  
		Neural Network's parameter set: $ N $, an Index: $ I $, Epochs: $ E $;
		\ENSURE ~~\\  
		Optimized parameters of $ N $;
		\STATE Randomly initialize each member of $ N $;
		\STATE for $ e $ in $ E $;
		\STATE \quad \quad for $ n $ in $ N $;
		\STATE \quad \quad  \quad \quad Make a random variation to $ n $. If it improves $ I $, keep it.  Otherwise, discard it;
		\RETURN $ N $.			
	\end{algorithmic}
\end{algorithm}

\subsection{Experiment \RNum{1}} 
In Experiment \RNum{1}, data 0, 1, and 2 are clustered with Neural Networks A, B, and C. We trained each Neural Network with 30 Epochs for each dataset. After training of the Neural Networks, we predict labels of 10,000 new data points, and plot the decision boundaries of each model. Figure \ref{fig:resdata0}, \ref{fig:resdata1}, and \ref{fig:resdata2} illustrate the clustering results and decision boundaries discovered by the Neural Networks, for each dataset, and compared with results of  K-means and Gaussian Mixture Model (GMM).  Note the interesting decision boundaries discovered by different Neural Networks for each dataset.

\subsection{Experiment \RNum{2}} 
In another paper \cite{liu2023min}, we introduced a new internal clustering evaluation index, called MMJ-based Silhouette coefficient (MMJ-SC). In Experiment \RNum{2}, we test CNNI with this new index. In the experiment, data 3 and 4 are clustered with Neural Network D, data 5 is clustered with Neural Network C. 

The result is shown in Figure \ref{fig:line_sep}. The last column of Figure \ref{fig:line_sep} plots the transformed data by the Neural Networks after convergence. To get a transformed point, after training of a Neural Network, we feed the Network with a data point, the values of the output neurons are the coordinate of the transformed data point. As shown by sub-figure (e) and (j) of Figure \ref{fig:line_sep}, linearly non-separable data becomes linearly separable by the line $ f(x) = x $, after transformation by the Neural Network.

Figure \ref{fig:other_clu} is clustering results of the two-moons and two-circles data by other models (from the scikit-learn project). It can be seen that only Spectral clustering, Agglomerative clustering, DBSCAN and OPTICS can deal with these non-convex shaped (non-flat geometry) data. However, all these models are non-parametric (transductive). That means we cannot get a set of parameters, and predict the label of a new point based on the parameters. Therefore, we can conclude that: CNNI equipped with MMJ-SC, is the first parametric (inductive) clustering model that can deal with non-convex shaped (non-flat geometry) data.

\section{Indices based on Fuzzy clustering}  
Fuzzy clustering (also referred to as soft clustering) is a form of clustering in which each data point can belong to more than one cluster with probability. As a comparison, in non-fuzzy clustering (also known as hard clustering), data are divided into distinct clusters, where each data point can only belong to exactly one cluster. By far we are using indices based on hard clustering, however, Neural Networks can naturally output soft clustering, by applying a Softmax activation to the last layer of a Neural Network.

In this section, we propose two soft clustering based indices and test one of them on two challenging data.

\subsection{Two soft clustering indices (SCIs)}  

\subsubsection{SCI V1} 
 
\begin{definition} 
	SCI V1
	
	\begin{equation}
	I_{sci} = \sum_{l\in  L } P(l)I(l)
	\label{equ:SCI_v1}
	\end{equation}
	
	where $ L $ is the set of all possible labeling of data $ X $; $ l $ belongs to $ L $; $ P(l) $ is the possibility of labeling $ l $, it is easy to compute when we have soft clustering of data $ X $; $ I(l) $ is an ordinary hard clustering based index, like SC or MMJ-SC.
\end{definition} 
SCI V1 is very expensive to calculate naively, since there are $ K^N $  possible labeling of data $ X $, where $ N $ is the number of points in $ X $, $ K $ is the number of clusters. Therefore, we do not test it in this paper. Hopefully mathematicians will find a smart way to calculate it efficiently.

\subsubsection{SCI V2}  
\begin{definition} 
	SCI V2
	\begin{equation}
	I  = \frac{1 }{N} \sum_{x\in  X } s(x)
	\label{equ:sc-ordinary}
	\end{equation}
	
	\begin{equation}
	I_{sci} = \frac{1 }{N} \sum_{x\in  X } P(x)s(x)
	\label{equ:SCI_v2}
	\end{equation}
	SCI V2 revises an ordinary hard clustering based index like SC or MMJ-SC, in which the index for a set of samples is given as the mean of index for each sample (Equation \ref{equ:sc-ordinary}).
	
	In the revision, we first determine the label for each sample $ x $, by finding the maximum probability of sample $ x $ belonging to a cluster; $ P(x) $ is the probability of sample $ x $ belonging to the cluster. After labels of all samples are determined, then we can calculate the index value $ s(x) $ for each sample $ x $. $ N $ is the number of points in data $ X $.
	
	If $ P(x)  = 1 $ for all samples $ x \in X $, then $ I_{sci} = I $.
\end{definition}

\subsection{A new optimization algorithm for CNNI} 
Naive coordinate descent (Algorithm 1) works good for simple clustering task, such as data 0-5 in Experiment \RNum{1} and \RNum{2}. For more challenging clustering tasks like data 6 and 7 in Figure \ref{fig:spiral}, naive coordinate descent seems not works. Therefore, a new optimization algorithm for dealing with these more challenging clustering tasks is designed.

The algorithm is called Step Back to See a Bigger Picture (SBSBP). In SBSBP, we set a criteria to identify a local minima. If the criteria for recognizing a  local minima is reached, we randomly step back some steps; making the Neural Network worse for clustering the data judged by the Index, so that the Neural Network can jump out of the local minima region, and try to step to the global minima. The best parameters so far are recorded so that we can restore it after the training.

A  local minima is identified if there is no improvement within some steps (Patience).  We could potentially accelerate the SBSBP algorithm by utilizing parallel computing, allowing for the simultaneous and random updating of multiple parameters in the Neural Network, rather than updating them one at a time.

\begin{algorithm}
	\caption{Step Back to See a Bigger Picture (SBSBP)}  
	\begin{algorithmic}[1]  
		\REQUIRE ~~\\  
		Neural Network's parameter set: $ N $, an Index: $ I $, Maximum steps: $ M $, hyper-parameter: $ Patience $;
		\ENSURE ~~\\  
		Optimized parameters of $ N $;
		\STATE Randomly initialize each member of $ N $;
		\STATE Calculate the Index $ I $;
		\STATE $ Counter \gets 0 $;
		\STATE for $ i $ in $ M $;
		\STATE \quad \quad Randomly select a parameter $ n $ from $ N $;
		\STATE \quad \quad Generate a random number $ \delta $;
		\STATE \quad \quad $ n \gets n + \delta $;
		\STATE \quad \quad Calculate the Index $ I $;
		\STATE \quad \quad If $ I $ improves:
		\STATE \quad \quad \quad \quad $ Counter \gets 0 $;
		\STATE \quad \quad \quad \quad Keep the variation to $ n $, continue;
		\STATE \quad \quad Elif $ I $ is invariant:
		\STATE \quad \quad \quad \quad $ Counter++ $;
		\STATE \quad \quad \quad \quad Restore previous value of $ n $, $ n \gets n - \delta $;
		\STATE \quad \quad Else:
		\STATE \quad \quad \quad \quad $ Counter++ $;
		\STATE \quad \quad \quad \quad If $ Counter > Patience $:
		\STATE \quad \quad \quad \quad  \quad \quad $ Counter \gets 0 $;
		\STATE \quad \quad \quad \quad  \quad \quad Randomly\_step\_back\_some\_steps;
		\STATE \quad \quad \quad \quad Else:
		\STATE \quad \quad \quad \quad  \quad \quad Restore previous value of $ n $, $ n \gets n - \delta $;

		\RETURN $ N $.			
	\end{algorithmic}
\end{algorithm}

\begin{algorithm}
	\caption{Randomly\_step\_back\_some\_steps}  
	\begin{algorithmic}[1]  
		\REQUIRE ~~\\  
		Neural Network's parameter set: $ N $, an Index: $ I $, hyper-parameter: $ \rho_1,  \rho_2, \xi$;
 
		\STATE $ counter2 \gets 0 $;
		\STATE $ infinite\_loop\_preventer \gets 0 $;
		\STATE  $ Steps \gets random.randint(\rho_1, \rho_2) $;  
		\STATE while $ counter2 <  Steps$ and $infinite\_loop\_preventer < \xi$;
		\STATE \quad \quad Randomly select a parameter $ n $ from $ N $;
		\STATE \quad \quad Generate a random number $ \delta $;
		\STATE \quad \quad $ n \gets n + \delta $;
		\STATE \quad \quad Calculate the Index $ I $;
		\STATE \quad \quad If $ I $ worsens:
		\STATE \quad \quad \quad \quad $ counter2++  $;
		\STATE \quad \quad \quad \quad Keep the variation to $ n $, continue;
		\STATE \quad \quad Else:
		\STATE \quad \quad \quad \quad Restore previous value of $ n $, $ n \gets n - \delta $;
		\STATE \quad \quad $infinite\_loop\_preventer ++$;
 		
	\end{algorithmic}
\end{algorithm}

\subsection{Experiment \RNum{3}} 
\begin{figure*} 
	\begin{subfigure}{0.23\textwidth}
		\includegraphics[width=\linewidth]{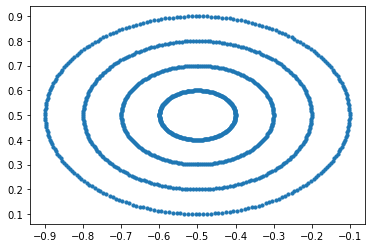}
		\caption{data 6}  
	\end{subfigure}    
	\begin{subfigure}{0.23\textwidth}
		\includegraphics[width=\linewidth]{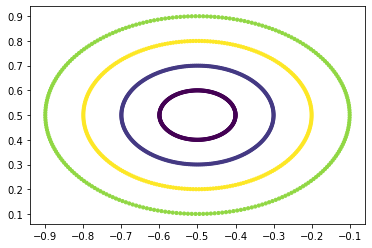}
		\caption{Clu-res}  
	\end{subfigure}    
	\begin{subfigure}{0.23\textwidth}
	\includegraphics[width=\linewidth]{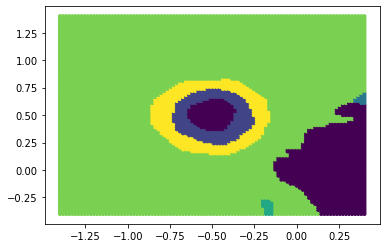}
	\caption{Deci-boun}  
\end{subfigure}    
\begin{subfigure}{0.23\textwidth}
\includegraphics[width=\linewidth]{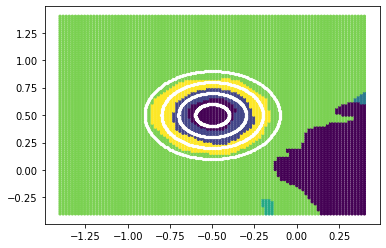}
\caption{Deci-boun-data}  
\end{subfigure}    
 
	\begin{subfigure}{0.23\textwidth}
	\includegraphics[width=\linewidth]{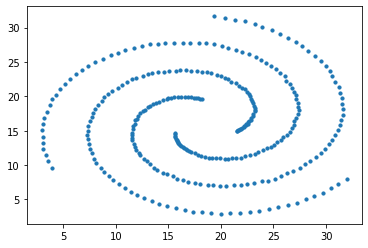}
	\caption{data 7}  
\end{subfigure}    
\begin{subfigure}{0.23\textwidth}
	\includegraphics[width=\linewidth]{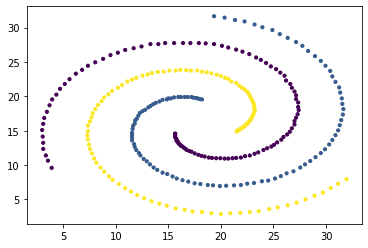}
	\caption{Clu-res}  
\end{subfigure}    
\begin{subfigure}{0.23\textwidth}
	\includegraphics[width=\linewidth]{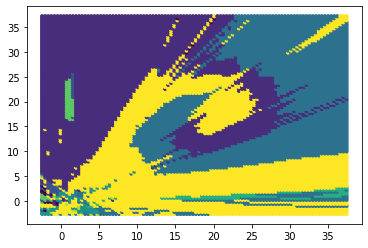}
	\caption{Deci-boun}  
\end{subfigure}    
\begin{subfigure}{0.23\textwidth}
	\includegraphics[width=\linewidth]{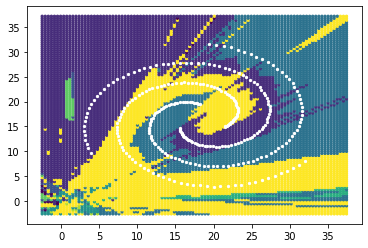}
	\caption{Deci-boun-data}  
\end{subfigure}    
 
	\caption{Clustering results and decision boundaries of data 6 and 7} \label{fig:spiral}	

\end{figure*}

In Experiment \RNum{3}, we revise MMJ-SC with SCI V2,  and tested it with more challenging clustering tasks of  data 6 and 7 in Figure \ref{fig:spiral}. Data 6 and 7 are referred to as Data 45 and Data 106 in \cite{liu2022new}. The data sources corresponding to the data IDs can be found at this URL. \footnote{\url{https://github.com/mike-liuliu/Min-Max-Jump-distance}}

The Neural Network used in Experiment \RNum{3} has similar structure as the Neural Networks in Figure \ref{fig:three-nn}, with one layer deeper and more nodes. It has shape  $2 \rightarrow 80\rightarrow 80 \rightarrow 80  \rightarrow 10$, which has 14,010 parameters. The activation to the last layer is changed to Softmax. As mentioned in Section \ref{sec:auto}, CNNI has some ability of automatic model selection. Therefore, we set the number of output neurons to 10 and let the Neural Network automatically select the true number of clusters. The Neural Network is optimized with SBSBP.

In the experiment, the $ Patience $ hyper-parameter is fixed. However, we can also make it dynamic, so that it is lager in later stages of training. When the index is a continuous value, we need to set a $ Tolerance $ hyper-parameter to decide whether the index has improved, or worsened, or unaffected. In the experiment, the $ Tolerance $ hyper-parameter is set to $ 10^{-8} $.

As shown by Figure \ref{fig:spiral}, after millions of steps of training, the Neural Network finally clustered the two data correctly, and learned reasonable (although unsatisfactory) decision boundaries.

\section{MMJ-K-means loss} 
Section \ref{sec:oth_loss} mentions we can use other losses like MMJ-K-means loss. In this section, we conduct experiments to test this idea.

Equation \ref{equ:k_means} is the loss of standard K-means.
\begin{equation}
\mathcal{L} = \frac{1 }{N} \sum\limits_{i=1}^N\sum\limits_{k=1}^K \mathds{1}\{c_i=k\} \lvert\lvert x_i-\mu_k\rvert\rvert^2 
\label{equ:k_means}
\end{equation}
where $ N $ is number of points in data $ X $, $ K $ is the number of clusters. $ \mu_k $ is the centroid of cluster $ k $, $ c_i $ is the label of data point $ x_i $.

In MMJ-K-means loss, Euclidean distance is replaced by MMJ distance \cite{liu2023min}, centroid of a cluster is replaced by the One-SCOM of the cluster \cite{liu2021topic,liu2023min}.

\subsection{Experiment \RNum{4}} 
\begin{figure} 
	\begin{subfigure}{0.23\textwidth}
		\includegraphics[width=\linewidth]{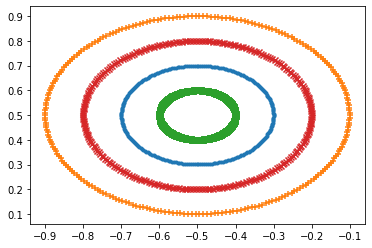}
		\caption{Clu-res}  
	\end{subfigure}    
	\begin{subfigure}{0.23\textwidth}
		\includegraphics[width=\linewidth]{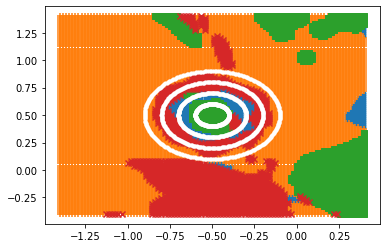}
		\caption{Deci-boun-data}  
	\end{subfigure}    
 
	\caption{Data 6 trained with MMJ-K-means loss} \label{fig:MMJKmeans}	

\end{figure}

In Experiment \RNum{4}, we tested MMJ-K-means loss with data 6. The Neural Network is optimized with SBSBP. Figure \ref{fig:MMJKmeans} illustrates the clustering result and decision boundary. The result shows MMJ-K-means loss also works for clustering, besides using an index as the loss function.

\subsection{MMJ-K-means loss based on Soft Clustering} 
We can also revise the MMJ-K-means loss to utilize Soft Clustering.
\begin{equation}
\mathcal{L}_s = \frac{1 }{N} \sum\limits_{i=1}^N\sum\limits_{k=1}^K \mathds{1}\{c_i=k\} \lvert\lvert x_i-\mu_k\rvert\rvert^2 (2 - P(x_i))
\label{equ:k_means_soft}
\end{equation}
Equation \ref{equ:k_means_soft} is similar to SCI V2. We first use Hard Clustering to decide which cluster data point $ x_i $ belongs to, then use Soft Clustering to decide the probability of data point $ x_i $ belongs to the cluster. 
$ P(x_i) $ is the probability. 

If $ P(x_i)  = 1 $ for all samples $ x_i \in X $, then $\mathcal{L}_s = \mathcal{L}$.

\subsection{Experiment \RNum{5}} 
In Experiment \RNum{5}, we tested MMJ-K-means loss based on Soft Clustering, with data 3 and 7. Section \ref{sec:mini-batch} discusses about training the Neural Network with Mini-Batches of data $ X $. In Experiment \RNum{5}, we use a variant of Mini-Batches, and devise a new optimization algorithm (Algorithm \ref{alg:sbsbp_mini}) for optimizing the Neural Network. In each iteration of Algorithm \ref{alg:sbsbp_mini}, we first draw a sample from data X, then train the Neural Network with the sample  and SBSBP for some steps, then check if the Neural Network has improved for clustering the whole data X. If it is, record the best parameters so far. The MMJ distance matrix of sample $ S $ is calculated under the context of data $ X $, not $ S $ itself. For the definition of \emph{context} in MMJ distance, see \cite{liu2023min}.
	\begin{figure*} 
	\begin{subfigure}{0.23\textwidth}
		\includegraphics[width=\linewidth]{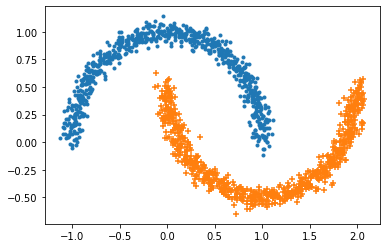}
		\caption{Clu-res}  
	\end{subfigure}    
	\begin{subfigure}{0.23\textwidth}
		\includegraphics[width=\linewidth]{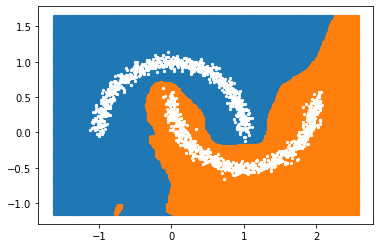}
		\caption{Deci-boun-data}  
	\end{subfigure}    
	\begin{subfigure}{0.23\textwidth}
	\includegraphics[width=\linewidth]{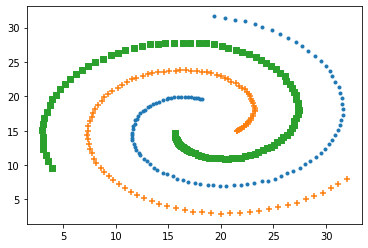}
	\caption{Clu-res}  
	\label{fig:over1}	
\end{subfigure}    
\begin{subfigure}{0.23\textwidth}
	\includegraphics[width=\linewidth]{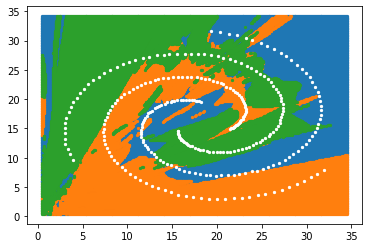}
	\caption{Deci-boun-data}  
	\label{fig:over2}	
\end{subfigure} 
 
	\caption{Data 3 and 7 trained with MMJ-K-means loss based on Soft Clustering (Equation \ref{equ:k_means_soft})} \label{fig:MMJKmeansSoft}	

\end{figure*}

\begin{algorithm}
	\caption{SBSBP and variant of Mini-Batch (MB-SBSBP)}  
	\label{alg:sbsbp_mini}
	\begin{algorithmic}[1]  
		\REQUIRE ~~\\  
		Neural Network's parameter set: $ N $, Loss: $ L $, Maximum iterations: $ T $, hyper-parameter: $ M $;
		\ENSURE ~~\\  
		Optimized parameters of $ N $;
		\STATE Randomly initialize each member of $ N $;
		\STATE for $ j $ in $ T $;
		\STATE \quad \quad Draw a sample $ S $ from data $ X $,  $S \subseteq X$;
		\STATE \quad \quad Train the Neural Network with sample $ S $ and SBSBP for $ M $ steps;
		\STATE \quad \quad Calculate the Loss $ L $ over the whole data $ X $;
		\STATE \quad \quad If $ L $ improves:
		\STATE \quad \quad \quad \quad Record parameter set $ N $;
 
		\RETURN Recorded best parameters so far $ N_{best} $.			
	\end{algorithmic}
\end{algorithm}

Figure \ref{fig:MMJKmeansSoft} illustrates the clustering results and decision boundaries. The result shows optimizing MMJ-K-means loss based on Soft Clustering also works for clustering.

The Neural Network used in Experiment \RNum{4} and  \RNum{5} has similar structure as the Neural Network used in Experiment \RNum{3}. It has shape  $2 \rightarrow 80\rightarrow 80 \rightarrow 80  \rightarrow N\_clusters$. We assume the true number of clusters (N\_clusters) is known in advance.

Section 6.2 of \cite{liu2023min} discusses the issue of  ``multiple One-SCOMs in one cluster", in Experiment \RNum{4} and \RNum{5}, we choose to keep them all. 

Note although K-means loss is used in this section, the K-means model itself is not used. Because a Neural Network can directly output Hard Clustering or Soft Clustering label of a data point, we do not need the K-means model to calculate the label of a data point. We just need the K-means loss.

\section{Discussion} 
\subsection{Other choices} 
Because the purpose of this paper is to introduce CNNI and test its  feasibility. We used simple fully connected feed-forward Neural Networks, with fixed activation function Tanh, and fixed clustering evaluation index Silhouette coefficient and MMJ-SC in the experiments. We did not test more advanced Neural Networks like convolutional neural network (CNN) or Bayesian neural network (BNN), and more advanced optimization techniques like dropout \cite{srivastava2014dropout}. 

For optimization algorithms, we also tested ``Nelder-Mead", ``Powell", ``CG", and ``BFGS" implemented by SciPy \cite{2020SciPy-NMeth}. It seems  ``Powell"  works good for CNNI. Other optimization algorithms like gradient descent, genetic algorithms, and other clustering evaluation indices were not tested either. 

Further research is necessary to make more elaborate selections of these factors. 

\subsection{Other losses}  \label{sec:oth_loss}
With labels of all data points known, another choice is to calculate other losses, such as the likelihood of Gaussian Mixture Model, or the loss of MMJ-K-means \cite{liu2023min}, then adjust Neural Network parameters by optimizing these losses.

\subsection{Model selection} 
Model selection  of CNNI is easy and natural. We test Neural Networks with different structures and different number of output neurons, and pick out the one which obtains the best internal-clustering-evaluation-index value for clustering the given dataset.
\subsubsection{Automatic model selection} \label{sec:auto}
It seems CNNI has some ability of automatic model selection. Suppose the true number of clusters is $ K $, we can set the number of output neurons to $ M $, where $ K  \ll  M $, hopefully, after training of the Neural Network, the number of active output neurons will degenerate to $ K $. We conducted an experiment to test automatic model selection. In the experiment, we use Neural Network D to cluster data 3 and 5. With the number of output neurons set to 30. The result is shown in Figure \ref{fig:model_selec}. It can be seen that in the region where data appears, the number of active output neurons degenerates to $ K $, with some minor error.

\subsection{Mini-Batch} \label{sec:mini-batch}
Computing the internal clustering evaluation index over the whole dataset may be time-consuming. We can divide the data into $ M $ Mini-Batches by random sampling, and use one Mini-Batch for training in one epoch. Or we can train $ M $ Neural Networks in parallel with the Mini-Batches, then calculate an average Neural Network, then train the average Neural Network with the whole dataset for several epochs.

\subsection{How to understand the learning?} 
In supervised learning, the learning of the Neural Network is guided by a tutor/teacher/supervisor, which is represented by a training dataset. 

How to understand the learning of CNNI? What is guiding the learning? Maybe we can consider the Neural Network is guided by some aesthetic sense, which is represented by an internal clustering evaluation index.

\subsection{Overfitting in unsupervised learning} 

	\begin{figure*} 
	\begin{subfigure}{0.3\textwidth}
	\includegraphics[width=\linewidth]{./images/v4imag/106-dis-centroid-clu-res-soft}
	\caption{Clu-res}  
\end{subfigure}  
	\begin{subfigure}{0.3\textwidth}
		\includegraphics[width=\linewidth]{./images/v4imag/106-dis-centroid-deci-boun-data-soft}
		\caption{Deci-boun-data}  
	\end{subfigure}    
	\begin{subfigure}{0.3\textwidth}
		\includegraphics[width=\linewidth]{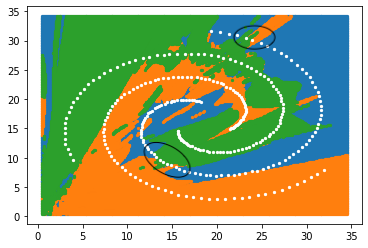}
		\caption{Overfitting}  
	\end{subfigure}    
	
	\caption{Overfitting in unsupervised learning?} \label{fig:Overfitting}	
	
\end{figure*}
Overfitting is common in supervised learning models. Can overfitting occur in unsupervised learning? Sub-figure \ref{fig:over1} and \ref{fig:over2} in Figure \ref{fig:MMJKmeansSoft}	 seems provide a new evidence, of overfitting in unsupervised learning.

As shown by Figure \ref{fig:Overfitting}, although the Neural Network labeled each training data point correctly, in the region indicated by the two black circles, the Neural Network failed to learn a reasonable decision boundary. It seems the Neural Network just ``memorized" the correct labels of training data points, and failed to generalize to new data points in the two regions.

\section{Conclusion and Future Works} 
We introduced a new clustering model called \emph{C}lustering with \emph{N}eural \emph{N}etwork and \emph{I}ndex (CNNI). Experiments are conducted to test the feasibility of CNNI. In the experiments, synthetic datasets were clustered by simple feed-forward Neural Networks (FNNs) with different model capacity. The result shows:

\begin{itemize}
	\item[$-$] CNNI can work properly for clustering data. 
	\item[$-$] Different structures of Neural Networks result in different decision boundaries.
	\item[$-$] CNNI equipped with MMJ-SC, achieves the first parametric (inductive) clustering model that can deal with non-convex shaped (non-flat geometry) data.
\end{itemize}

 Further research is necessary to test other types of Neural Networks, optimization techniques, and other clustering evaluation indices. And how CNNI performs in higher dimensions.
 
A potential advantage of CNNI is its flexibility of model capacity. When the hyperparameter $ K $ (number of clusters) is fixed, the model capacity (number of learnable parameters) of other parametric clustering models like K-means and Gaussian Mixture Model (GMM) is fixed.  The model capacity of CNNI is flexible due to the utilization of Neural Network.
 
\bibliographystyle{ACM-Reference-Format}
\bibliography{new_index} 


\begin{thebibliography}{38}


\ifx \showCODEN    \undefined \def \showCODEN     #1{\unskip}     \fi
\ifx \showDOI      \undefined \def \showDOI       #1{#1}\fi
\ifx \showISBNx    \undefined \def \showISBNx     #1{\unskip}     \fi
\ifx \showISBNxiii \undefined \def \showISBNxiii  #1{\unskip}     \fi
\ifx \showISSN     \undefined \def \showISSN      #1{\unskip}     \fi
\ifx \showLCCN     \undefined \def \showLCCN      #1{\unskip}     \fi
\ifx \shownote     \undefined \def \shownote      #1{#1}          \fi
\ifx \showarticletitle \undefined \def \showarticletitle #1{#1}   \fi
\ifx \showURL      \undefined \def \showURL       {\relax}        \fi
\providecommand\bibfield[2]{#2}
\providecommand\bibinfo[2]{#2}
\providecommand\natexlab[1]{#1}
\providecommand\showeprint[2][]{arXiv:#2}

\bibitem[\protect\citeauthoryear{Aljalbout, Golkov, Siddiqui, Strobel, and
  Cremers}{Aljalbout et~al\mbox{.}}{2018}]%
        {aljalbout2018clustering}
\bibfield{author}{\bibinfo{person}{Elie Aljalbout}, \bibinfo{person}{Vladimir
  Golkov}, \bibinfo{person}{Yawar Siddiqui}, \bibinfo{person}{Maximilian
  Strobel}, {and} \bibinfo{person}{Daniel Cremers}.}
  \bibinfo{year}{2018}\natexlab{}.
\newblock \showarticletitle{Clustering with deep learning: Taxonomy and new
  methods}.
\newblock \bibinfo{journal}{\emph{arXiv preprint arXiv:1801.07648}}
  (\bibinfo{year}{2018}).
\newblock


\bibitem[\protect\citeauthoryear{Anderson}{Anderson}{1995}]%
        {anderson1995introduction}
\bibfield{author}{\bibinfo{person}{James~A Anderson}.}
  \bibinfo{year}{1995}\natexlab{}.
\newblock \bibinfo{booktitle}{\emph{An introduction to neural networks}}.
\newblock \bibinfo{publisher}{MIT press}.
\newblock


\bibitem[\protect\citeauthoryear{Aranganayagi and Thangavel}{Aranganayagi and
  Thangavel}{2007}]%
        {aranganayagi2007clustering}
\bibfield{author}{\bibinfo{person}{S Aranganayagi} {and}
  \bibinfo{person}{Kuttiyannan Thangavel}.} \bibinfo{year}{2007}\natexlab{}.
\newblock \showarticletitle{Clustering categorical data using silhouette
  coefficient as a relocating measure}. In
  \bibinfo{booktitle}{\emph{International conference on computational
  intelligence and multimedia applications (ICCIMA 2007)}},
  Vol.~\bibinfo{volume}{2}. IEEE, \bibinfo{pages}{13--17}.
\newblock


\bibitem[\protect\citeauthoryear{Baya and Granitto}{Baya and Granitto}{2013}]%
        {baya2013many}
\bibfield{author}{\bibinfo{person}{Ariel~E Baya} {and} \bibinfo{person}{Pablo~M
  Granitto}.} \bibinfo{year}{2013}\natexlab{}.
\newblock \showarticletitle{How many clusters: A validation index for
  arbitrary-shaped clusters}.
\newblock \bibinfo{journal}{\emph{IEEE/ACM Transactions on Computational
  Biology and Bioinformatics}} \bibinfo{volume}{10}, \bibinfo{number}{2}
  (\bibinfo{year}{2013}), \bibinfo{pages}{401--414}.
\newblock


\bibitem[\protect\citeauthoryear{Bezdek and Pal}{Bezdek and Pal}{1995}]%
        {bezdek1995cluster}
\bibfield{author}{\bibinfo{person}{James~C Bezdek} {and}
  \bibinfo{person}{Nikhil~R Pal}.} \bibinfo{year}{1995}\natexlab{}.
\newblock \showarticletitle{Cluster validation with generalized Dunn's
  indices}. In \bibinfo{booktitle}{\emph{Proceedings 1995 second New Zealand
  international two-stream conference on artificial neural networks and expert
  systems}}. IEEE Computer Society, \bibinfo{pages}{190--190}.
\newblock


\bibitem[\protect\citeauthoryear{Chou, Su, and Lai}{Chou et~al\mbox{.}}{2004}]%
        {chou2004new}
\bibfield{author}{\bibinfo{person}{C-H Chou}, \bibinfo{person}{M-C Su}, {and}
  \bibinfo{person}{Eugene Lai}.} \bibinfo{year}{2004}\natexlab{}.
\newblock \showarticletitle{A new cluster validity measure and its application
  to image compression}.
\newblock \bibinfo{journal}{\emph{Pattern Analysis and Applications}}
  \bibinfo{volume}{7}, \bibinfo{number}{2} (\bibinfo{year}{2004}),
  \bibinfo{pages}{205--220}.
\newblock


\bibitem[\protect\citeauthoryear{Guan and Loew}{Guan and Loew}{2020}]%
        {guan2020internal}
\bibfield{author}{\bibinfo{person}{Shuyue Guan} {and} \bibinfo{person}{Murray
  Loew}.} \bibinfo{year}{2020}\natexlab{}.
\newblock \showarticletitle{An internal cluster validity index using a
  distance-based separability measure}. In \bibinfo{booktitle}{\emph{2020 IEEE
  32nd international conference on tools with artificial intelligence
  (ICTAI)}}. IEEE, \bibinfo{pages}{827--834}.
\newblock


\bibitem[\protect\citeauthoryear{Gurney}{Gurney}{2018}]%
        {gurney2018introduction}
\bibfield{author}{\bibinfo{person}{Kevin Gurney}.}
  \bibinfo{year}{2018}\natexlab{}.
\newblock \bibinfo{booktitle}{\emph{An introduction to neural networks}}.
\newblock \bibinfo{publisher}{CRC press}.
\newblock


\bibitem[\protect\citeauthoryear{Halkidi, Batistakis, and Vazirgiannis}{Halkidi
  et~al\mbox{.}}{2001}]%
        {halkidi2001clustering2001}
\bibfield{author}{\bibinfo{person}{Maria Halkidi}, \bibinfo{person}{Yannis
  Batistakis}, {and} \bibinfo{person}{Michalis Vazirgiannis}.}
  \bibinfo{year}{2001}\natexlab{}.
\newblock \showarticletitle{On clustering validation techniques}.
\newblock \bibinfo{journal}{\emph{Journal of intelligent information systems}}
  \bibinfo{volume}{17}, \bibinfo{number}{2} (\bibinfo{year}{2001}),
  \bibinfo{pages}{107--145}.
\newblock


\bibitem[\protect\citeauthoryear{Halkidi and Vazirgiannis}{Halkidi and
  Vazirgiannis}{2001}]%
        {halkidi2001clustering}
\bibfield{author}{\bibinfo{person}{Maria Halkidi} {and}
  \bibinfo{person}{Michalis Vazirgiannis}.} \bibinfo{year}{2001}\natexlab{}.
\newblock \showarticletitle{Clustering validity assessment: Finding the optimal
  partitioning of a data set}. In \bibinfo{booktitle}{\emph{Proceedings 2001
  IEEE international conference on data mining}}. IEEE,
  \bibinfo{pages}{187--194}.
\newblock


\bibitem[\protect\citeauthoryear{Halkidi and Vazirgiannis}{Halkidi and
  Vazirgiannis}{2008}]%
        {halkidi2008density}
\bibfield{author}{\bibinfo{person}{Maria Halkidi} {and}
  \bibinfo{person}{Michalis Vazirgiannis}.} \bibinfo{year}{2008}\natexlab{}.
\newblock \showarticletitle{A density-based cluster validity approach using
  multi-representatives}.
\newblock \bibinfo{journal}{\emph{Pattern Recognition Letters}}
  \bibinfo{volume}{29}, \bibinfo{number}{6} (\bibinfo{year}{2008}),
  \bibinfo{pages}{773--786}.
\newblock


\bibitem[\protect\citeauthoryear{Hu and Zhong}{Hu and Zhong}{2019}]%
        {hu2019internal}
\bibfield{author}{\bibinfo{person}{Lianyu Hu} {and} \bibinfo{person}{Caiming
  Zhong}.} \bibinfo{year}{2019}\natexlab{}.
\newblock \showarticletitle{An internal validity index based on
  density-involved distance}.
\newblock \bibinfo{journal}{\emph{IEEE Access}}  \bibinfo{volume}{7}
  (\bibinfo{year}{2019}), \bibinfo{pages}{40038--40051}.
\newblock


\bibitem[\protect\citeauthoryear{Johnson}{Johnson}{1967}]%
        {johnson1967hierarchical}
\bibfield{author}{\bibinfo{person}{Stephen~C Johnson}.}
  \bibinfo{year}{1967}\natexlab{}.
\newblock \showarticletitle{Hierarchical clustering schemes}.
\newblock \bibinfo{journal}{\emph{Psychometrika}} \bibinfo{volume}{32},
  \bibinfo{number}{3} (\bibinfo{year}{1967}), \bibinfo{pages}{241--254}.
\newblock


\bibitem[\protect\citeauthoryear{Kohonen}{Kohonen}{1990}]%
        {kohonen1990self}
\bibfield{author}{\bibinfo{person}{Teuvo Kohonen}.}
  \bibinfo{year}{1990}\natexlab{}.
\newblock \showarticletitle{The self-organizing map}.
\newblock \bibinfo{journal}{\emph{Proc. IEEE}} \bibinfo{volume}{78},
  \bibinfo{number}{9} (\bibinfo{year}{1990}), \bibinfo{pages}{1464--1480}.
\newblock


\bibitem[\protect\citeauthoryear{Li, Yue, Wang, Ding, and Li}{Li
  et~al\mbox{.}}{2020}]%
        {li2020new}
\bibfield{author}{\bibinfo{person}{Qi Li}, \bibinfo{person}{Shihong Yue},
  \bibinfo{person}{Yaru Wang}, \bibinfo{person}{Mingliang Ding}, {and}
  \bibinfo{person}{Jia Li}.} \bibinfo{year}{2020}\natexlab{}.
\newblock \showarticletitle{A new cluster validity index based on the
  adjustment of within-cluster distance}.
\newblock \bibinfo{journal}{\emph{IEEE Access}}  \bibinfo{volume}{8}
  (\bibinfo{year}{2020}), \bibinfo{pages}{202872--202885}.
\newblock


\bibitem[\protect\citeauthoryear{Liu}{Liu}{2021}]%
        {liu2021topic}
\bibfield{author}{\bibinfo{person}{Gangli Liu}.}
  \bibinfo{year}{2021}\natexlab{}.
\newblock \showarticletitle{Topic model supervised by understanding map}.
\newblock \bibinfo{journal}{\emph{arXiv preprint arXiv:2110.06043}}
  (\bibinfo{year}{2021}).
\newblock


\bibitem[\protect\citeauthoryear{Liu}{Liu}{2022}]%
        {liu2022new}
\bibfield{author}{\bibinfo{person}{Gangli Liu}.}
  \bibinfo{year}{2022}\natexlab{}.
\newblock \showarticletitle{A New Index for Clustering Evaluation Based on
  Density Estimation}.
\newblock \bibinfo{journal}{\emph{arXiv preprint arXiv:2207.01294}}
  (\bibinfo{year}{2022}).
\newblock


\bibitem[\protect\citeauthoryear{Liu}{Liu}{2023}]%
        {liu2023min}
\bibfield{author}{\bibinfo{person}{Gangli Liu}.}
  \bibinfo{year}{2023}\natexlab{}.
\newblock \showarticletitle{Min-Max-Jump distance and its applications}.
\newblock \bibinfo{journal}{\emph{arXiv preprint arXiv:2301.05994}}
  (\bibinfo{year}{2023}).
\newblock


\bibitem[\protect\citeauthoryear{Liu, Li, Xiong, Gao, Wu, and Wu}{Liu
  et~al\mbox{.}}{2013}]%
        {liu2013understanding}
\bibfield{author}{\bibinfo{person}{Yanchi Liu}, \bibinfo{person}{Zhongmou Li},
  \bibinfo{person}{Hui Xiong}, \bibinfo{person}{Xuedong Gao},
  \bibinfo{person}{Junjie Wu}, {and} \bibinfo{person}{Sen Wu}.}
  \bibinfo{year}{2013}\natexlab{}.
\newblock \showarticletitle{Understanding and enhancement of internal
  clustering validation measures}.
\newblock \bibinfo{journal}{\emph{IEEE transactions on cybernetics}}
  \bibinfo{volume}{43}, \bibinfo{number}{3} (\bibinfo{year}{2013}),
  \bibinfo{pages}{982--994}.
\newblock


\bibitem[\protect\citeauthoryear{Lu}{Lu}{2022}]%
        {lu2022learning}
\bibfield{author}{\bibinfo{person}{Wei Lu}.} \bibinfo{year}{2022}\natexlab{}.
\newblock \emph{\bibinfo{title}{Learning Guarantees for Graph Convolutional
  Networks in the Stochastic Block Model}}.
\newblock \bibinfo{thesistype}{Ph.\,D. Dissertation}. \bibinfo{school}{Brandeis
  University}.
\newblock


\bibitem[\protect\citeauthoryear{Maulik and Bandyopadhyay}{Maulik and
  Bandyopadhyay}{2002}]%
        {maulik2002performance}
\bibfield{author}{\bibinfo{person}{Ujjwal Maulik} {and}
  \bibinfo{person}{Sanghamitra Bandyopadhyay}.}
  \bibinfo{year}{2002}\natexlab{}.
\newblock \showarticletitle{Performance evaluation of some clustering
  algorithms and validity indices}.
\newblock \bibinfo{journal}{\emph{IEEE Transactions on pattern analysis and
  machine intelligence}} \bibinfo{volume}{24}, \bibinfo{number}{12}
  (\bibinfo{year}{2002}), \bibinfo{pages}{1650--1654}.
\newblock


\bibitem[\protect\citeauthoryear{Moulavi, Jaskowiak, Campello, Zimek, and
  Sander}{Moulavi et~al\mbox{.}}{2014}]%
        {moulavi2014density}
\bibfield{author}{\bibinfo{person}{Davoud Moulavi}, \bibinfo{person}{Pablo~A
  Jaskowiak}, \bibinfo{person}{Ricardo~JGB Campello}, \bibinfo{person}{Arthur
  Zimek}, {and} \bibinfo{person}{J{\"o}rg Sander}.}
  \bibinfo{year}{2014}\natexlab{}.
\newblock \showarticletitle{Density-based clustering validation}. In
  \bibinfo{booktitle}{\emph{Proceedings of the 2014 SIAM international
  conference on data mining}}. SIAM, \bibinfo{pages}{839--847}.
\newblock


\bibitem[\protect\citeauthoryear{Ng, Jordan, and Weiss}{Ng
  et~al\mbox{.}}{2001}]%
        {ng2001spectral}
\bibfield{author}{\bibinfo{person}{Andrew Ng}, \bibinfo{person}{Michael
  Jordan}, {and} \bibinfo{person}{Yair Weiss}.}
  \bibinfo{year}{2001}\natexlab{}.
\newblock \showarticletitle{On spectral clustering: Analysis and an algorithm}.
\newblock \bibinfo{journal}{\emph{Advances in neural information processing
  systems}}  \bibinfo{volume}{14} (\bibinfo{year}{2001}).
\newblock


\bibitem[\protect\citeauthoryear{Pedregosa, Varoquaux, Gramfort, Michel,
  Thirion, Grisel, Blondel, Prettenhofer, Weiss, Dubourg, Vanderplas, Passos,
  Cournapeau, Brucher, Perrot, and Duchesnay}{Pedregosa et~al\mbox{.}}{2011}]%
        {scikit-learn}
\bibfield{author}{\bibinfo{person}{F. Pedregosa}, \bibinfo{person}{G.
  Varoquaux}, \bibinfo{person}{A. Gramfort}, \bibinfo{person}{V. Michel},
  \bibinfo{person}{B. Thirion}, \bibinfo{person}{O. Grisel},
  \bibinfo{person}{M. Blondel}, \bibinfo{person}{P. Prettenhofer},
  \bibinfo{person}{R. Weiss}, \bibinfo{person}{V. Dubourg}, \bibinfo{person}{J.
  Vanderplas}, \bibinfo{person}{A. Passos}, \bibinfo{person}{D. Cournapeau},
  \bibinfo{person}{M. Brucher}, \bibinfo{person}{M. Perrot}, {and}
  \bibinfo{person}{E. Duchesnay}.} \bibinfo{year}{2011}\natexlab{}.
\newblock \showarticletitle{Scikit-learn: Machine Learning in {P}ython}.
\newblock \bibinfo{journal}{\emph{Journal of Machine Learning Research}}
  \bibinfo{volume}{12} (\bibinfo{year}{2011}), \bibinfo{pages}{2825--2830}.
\newblock


\bibitem[\protect\citeauthoryear{Petrovi{\'c}}{Petrovi{\'c}}{2006}]%
        {petrovic2006comparison}
\bibfield{author}{\bibinfo{person}{Slobodan Petrovi{\'c}}.}
  \bibinfo{year}{2006}\natexlab{}.
\newblock \showarticletitle{A comparison between the silhouette index and the
  davies-bouldin index in labelling ids clusters}.
\newblock  (\bibinfo{year}{2006}).
\newblock


\bibitem[\protect\citeauthoryear{Pfitzner, Leibbrandt, and Powers}{Pfitzner
  et~al\mbox{.}}{2009}]%
        {pfitzner2009characterization}
\bibfield{author}{\bibinfo{person}{Darius Pfitzner}, \bibinfo{person}{Richard
  Leibbrandt}, {and} \bibinfo{person}{David Powers}.}
  \bibinfo{year}{2009}\natexlab{}.
\newblock \showarticletitle{Characterization and evaluation of similarity
  measures for pairs of clusterings}.
\newblock \bibinfo{journal}{\emph{Knowledge and Information Systems}}
  \bibinfo{volume}{19}, \bibinfo{number}{3} (\bibinfo{year}{2009}),
  \bibinfo{pages}{361--394}.
\newblock


\bibitem[\protect\citeauthoryear{Ren, Pu, Yang, Xu, Li, Pu, Yu, and He}{Ren
  et~al\mbox{.}}{2022}]%
        {ren2022deep}
\bibfield{author}{\bibinfo{person}{Yazhou Ren}, \bibinfo{person}{Jingyu Pu},
  \bibinfo{person}{Zhimeng Yang}, \bibinfo{person}{Jie Xu},
  \bibinfo{person}{Guofeng Li}, \bibinfo{person}{Xiaorong Pu},
  \bibinfo{person}{Philip~S Yu}, {and} \bibinfo{person}{Lifang He}.}
  \bibinfo{year}{2022}\natexlab{}.
\newblock \showarticletitle{Deep clustering: A comprehensive survey}.
\newblock \bibinfo{journal}{\emph{arXiv preprint arXiv:2210.04142}}
  (\bibinfo{year}{2022}).
\newblock


\bibitem[\protect\citeauthoryear{Schubert, Sander, Ester, Kriegel, and
  Xu}{Schubert et~al\mbox{.}}{2017}]%
        {schubert2017dbscan}
\bibfield{author}{\bibinfo{person}{Erich Schubert}, \bibinfo{person}{J{\"o}rg
  Sander}, \bibinfo{person}{Martin Ester}, \bibinfo{person}{Hans~Peter
  Kriegel}, {and} \bibinfo{person}{Xiaowei Xu}.}
  \bibinfo{year}{2017}\natexlab{}.
\newblock \showarticletitle{DBSCAN revisited, revisited: why and how you should
  (still) use DBSCAN}.
\newblock \bibinfo{journal}{\emph{ACM Transactions on Database Systems (TODS)}}
  \bibinfo{volume}{42}, \bibinfo{number}{3} (\bibinfo{year}{2017}),
  \bibinfo{pages}{1--21}.
\newblock


\bibitem[\protect\citeauthoryear{{\c{S}}enol}{{\c{S}}enol}{2022}]%
        {csenol2022viasckde}
\bibfield{author}{\bibinfo{person}{Ali {\c{S}}enol}.}
  \bibinfo{year}{2022}\natexlab{}.
\newblock \showarticletitle{VIASCKDE Index: A Novel Internal Cluster Validity
  Index for Arbitrary-Shaped Clusters Based on the Kernel Density Estimation}.
\newblock \bibinfo{journal}{\emph{Computational Intelligence and Neuroscience}}
   \bibinfo{volume}{2022} (\bibinfo{year}{2022}).
\newblock


\bibitem[\protect\citeauthoryear{Srivastava, Hinton, Krizhevsky, Sutskever, and
  Salakhutdinov}{Srivastava et~al\mbox{.}}{2014}]%
        {srivastava2014dropout}
\bibfield{author}{\bibinfo{person}{Nitish Srivastava},
  \bibinfo{person}{Geoffrey Hinton}, \bibinfo{person}{Alex Krizhevsky},
  \bibinfo{person}{Ilya Sutskever}, {and} \bibinfo{person}{Ruslan
  Salakhutdinov}.} \bibinfo{year}{2014}\natexlab{}.
\newblock \showarticletitle{Dropout: a simple way to prevent neural networks
  from overfitting}.
\newblock \bibinfo{journal}{\emph{The journal of machine learning research}}
  \bibinfo{volume}{15}, \bibinfo{number}{1} (\bibinfo{year}{2014}),
  \bibinfo{pages}{1929--1958}.
\newblock


\bibitem[\protect\citeauthoryear{Vendramin, Campello, and Hruschka}{Vendramin
  et~al\mbox{.}}{2010}]%
        {vendramin2010relative}
\bibfield{author}{\bibinfo{person}{Lucas Vendramin},
  \bibinfo{person}{Ricardo~JGB Campello}, {and} \bibinfo{person}{Eduardo~R
  Hruschka}.} \bibinfo{year}{2010}\natexlab{}.
\newblock \showarticletitle{Relative clustering validity criteria: A
  comparative overview}.
\newblock \bibinfo{journal}{\emph{Statistical analysis and data mining: the ASA
  data science journal}} \bibinfo{volume}{3}, \bibinfo{number}{4}
  (\bibinfo{year}{2010}), \bibinfo{pages}{209--235}.
\newblock


\bibitem[\protect\citeauthoryear{Virtanen, Gommers, Oliphant, Haberland, Reddy,
  Cournapeau, Burovski, Peterson, Weckesser, Bright, {van der Walt}, Brett,
  Wilson, Millman, Mayorov, Nelson, Jones, Kern, Larson, Carey, Polat, Feng,
  Moore, {VanderPlas}, Laxalde, Perktold, Cimrman, Henriksen, Quintero, Harris,
  Archibald, Ribeiro, Pedregosa, {van Mulbregt}, and {SciPy 1.0
  Contributors}}{Virtanen et~al\mbox{.}}{2020}]%
        {2020SciPy-NMeth}
\bibfield{author}{\bibinfo{person}{Pauli Virtanen}, \bibinfo{person}{Ralf
  Gommers}, \bibinfo{person}{Travis~E. Oliphant}, \bibinfo{person}{Matt
  Haberland}, \bibinfo{person}{Tyler Reddy}, \bibinfo{person}{David
  Cournapeau}, \bibinfo{person}{Evgeni Burovski}, \bibinfo{person}{Pearu
  Peterson}, \bibinfo{person}{Warren Weckesser}, \bibinfo{person}{Jonathan
  Bright}, \bibinfo{person}{St{\'e}fan~J. {van der Walt}},
  \bibinfo{person}{Matthew Brett}, \bibinfo{person}{Joshua Wilson},
  \bibinfo{person}{K.~Jarrod Millman}, \bibinfo{person}{Nikolay Mayorov},
  \bibinfo{person}{Andrew R.~J. Nelson}, \bibinfo{person}{Eric Jones},
  \bibinfo{person}{Robert Kern}, \bibinfo{person}{Eric Larson},
  \bibinfo{person}{C~J Carey}, \bibinfo{person}{{\.I}lhan Polat},
  \bibinfo{person}{Yu Feng}, \bibinfo{person}{Eric~W. Moore},
  \bibinfo{person}{Jake {VanderPlas}}, \bibinfo{person}{Denis Laxalde},
  \bibinfo{person}{Josef Perktold}, \bibinfo{person}{Robert Cimrman},
  \bibinfo{person}{Ian Henriksen}, \bibinfo{person}{E.~A. Quintero},
  \bibinfo{person}{Charles~R. Harris}, \bibinfo{person}{Anne~M. Archibald},
  \bibinfo{person}{Ant{\^o}nio~H. Ribeiro}, \bibinfo{person}{Fabian Pedregosa},
  \bibinfo{person}{Paul {van Mulbregt}}, {and} \bibinfo{person}{{SciPy 1.0
  Contributors}}.} \bibinfo{year}{2020}\natexlab{}.
\newblock \showarticletitle{{{SciPy} 1.0: Fundamental Algorithms for Scientific
  Computing in Python}}.
\newblock \bibinfo{journal}{\emph{Nature Methods}}  \bibinfo{volume}{17}
  (\bibinfo{year}{2020}), \bibinfo{pages}{261--272}.
\newblock
\urldef\tempurl%
\url{https://doi.org/10.1038/s41592-019-0686-2}
\showDOI{\tempurl}


\bibitem[\protect\citeauthoryear{Xie and Beni}{Xie and Beni}{1991}]%
        {xie1991validity}
\bibfield{author}{\bibinfo{person}{Xuanli~Lisa Xie} {and}
  \bibinfo{person}{Gerardo Beni}.} \bibinfo{year}{1991}\natexlab{}.
\newblock \showarticletitle{A validity measure for fuzzy clustering}.
\newblock \bibinfo{journal}{\emph{IEEE Transactions on pattern analysis and
  machine intelligence}} \bibinfo{volume}{13}, \bibinfo{number}{8}
  (\bibinfo{year}{1991}), \bibinfo{pages}{841--847}.
\newblock


\bibitem[\protect\citeauthoryear{Xu, Zhang, Liu, and Luo}{Xu
  et~al\mbox{.}}{2020}]%
        {xu2020efficient}
\bibfield{author}{\bibinfo{person}{Qin Xu}, \bibinfo{person}{Qiang Zhang},
  \bibinfo{person}{Jinpei Liu}, {and} \bibinfo{person}{Bin Luo}.}
  \bibinfo{year}{2020}\natexlab{}.
\newblock \showarticletitle{Efficient synthetical clustering validity indexes
  for hierarchical clustering}.
\newblock \bibinfo{journal}{\emph{Expert Systems with Applications}}
  \bibinfo{volume}{151} (\bibinfo{year}{2020}), \bibinfo{pages}{113367}.
\newblock


\bibitem[\protect\citeauthoryear{Yeung and Ruzzo}{Yeung and Ruzzo}{2001}]%
        {yeung2001details}
\bibfield{author}{\bibinfo{person}{Ka~Yee Yeung} {and}
  \bibinfo{person}{Walter~L Ruzzo}.} \bibinfo{year}{2001}\natexlab{}.
\newblock \showarticletitle{Details of the adjusted rand index and clustering
  algorithms, supplement to the paper an empirical study on principal component
  analysis for clustering gene expression data}.
\newblock \bibinfo{journal}{\emph{Bioinformatics}} \bibinfo{volume}{17},
  \bibinfo{number}{9} (\bibinfo{year}{2001}), \bibinfo{pages}{763--774}.
\newblock


\bibitem[\protect\citeauthoryear{{\v{Z}}alik and {\v{Z}}alik}{{\v{Z}}alik and
  {\v{Z}}alik}{2011}]%
        {vzalik2011validity}
\bibfield{author}{\bibinfo{person}{Krista~Rizman {\v{Z}}alik} {and}
  \bibinfo{person}{Borut {\v{Z}}alik}.} \bibinfo{year}{2011}\natexlab{}.
\newblock \showarticletitle{Validity index for clusters of different sizes and
  densities}.
\newblock \bibinfo{journal}{\emph{Pattern Recognition Letters}}
  \bibinfo{volume}{32}, \bibinfo{number}{2} (\bibinfo{year}{2011}),
  \bibinfo{pages}{221--234}.
\newblock


\bibitem[\protect\citeauthoryear{Zhang, Ramakrishnan, and Livny}{Zhang
  et~al\mbox{.}}{1996}]%
        {zhang1996birch}
\bibfield{author}{\bibinfo{person}{Tian Zhang}, \bibinfo{person}{Raghu
  Ramakrishnan}, {and} \bibinfo{person}{Miron Livny}.}
  \bibinfo{year}{1996}\natexlab{}.
\newblock \showarticletitle{BIRCH: an efficient data clustering method for very
  large databases}.
\newblock \bibinfo{journal}{\emph{ACM sigmod record}} \bibinfo{volume}{25},
  \bibinfo{number}{2} (\bibinfo{year}{1996}), \bibinfo{pages}{103--114}.
\newblock


\bibitem[\protect\citeauthoryear{Zhou, Xu, Zheng, Chen, Bu, Wu, Wang, Zhu,
  Ester, et~al\mbox{.}}{Zhou et~al\mbox{.}}{2022}]%
        {zhou2022comprehensive}
\bibfield{author}{\bibinfo{person}{Sheng Zhou}, \bibinfo{person}{Hongjia Xu},
  \bibinfo{person}{Zhuonan Zheng}, \bibinfo{person}{Jiawei Chen},
  \bibinfo{person}{Jiajun Bu}, \bibinfo{person}{Jia Wu}, \bibinfo{person}{Xin
  Wang}, \bibinfo{person}{Wenwu Zhu}, \bibinfo{person}{Martin Ester},
  {et~al\mbox{.}}} \bibinfo{year}{2022}\natexlab{}.
\newblock \showarticletitle{A comprehensive survey on deep clustering:
  Taxonomy, challenges, and future directions}.
\newblock \bibinfo{journal}{\emph{arXiv preprint arXiv:2206.07579}}
  (\bibinfo{year}{2022}).
\newblock


\end{thebibliography}
 
\end{document}